\documentclass[letterpaper, 10 pt, conference]{ieeeconf}

\IEEEoverridecommandlockouts
\usepackage[utf8]{inputenc}
\usepackage[T1]{fontenc}
\usepackage{cite}
\usepackage{amsmath,amssymb,amsfonts}
\usepackage{algorithmic}
\usepackage{graphicx}
\usepackage{textcomp}
\usepackage{xcolor}
\usepackage{hyperref}
\usepackage{subfig}
\usepackage{amssymb,mathtools,bm}
\usepackage{booktabs}
\usepackage{pifont}
\usepackage{comment}
\usepackage{academicons}
\usepackage{tikz}
\usepackage{stfloats}
\usepackage{multirow}
\renewcommand{\baselinestretch}{0.96}
\usepackage[compact]{titlesec}
\usetikzlibrary{shapes.geometric, arrows.meta, positioning, fit, backgrounds, calc}
\definecolor{orcidlogocol}{HTML}{A6CE39}
\def\BibTeX{{\rm B\kern-.05em{\sc i\kern-.025em b}\kern-.08em
    T\kern-.1667em\lower.7ex\hbox{E}\kern-.125emX}}

\usepackage{scalerel}
\usetikzlibrary{svg.path}

\tikzset{
  orcidlogo/.pic={
    \fill[orcidlogocol] svg{M256,128c0,70.7-57.3,128-128,128C57.3,256,0,198.7,0,128C0,57.3,57.3,0,128,0C198.7,0,256,57.3,256,128z};
    \fill[white] svg{M86.3,186.2H70.9V79.1h15.4v48.4V186.2z}
                 svg{M108.9,79.1h41.6c39.6,0,57,28.3,57,53.6c0,27.5-21.5,53.6-56.8,53.6h-41.8V79.1z M124.3,172.4h24.5c34.9,0,42.9-26.5,42.9-39.7c0-21.5-13.7-39.7-43.7-39.7h-23.7V172.4z}
                 svg{M88.7,56.8c0,5.5-4.5,10.1-10.1,10.1c-5.6,0-10.1-4.6-10.1-10.1c0-5.6,4.5-10.1,10.1-10.1C84.2,46.7,88.7,51.3,88.7,56.8z};
  }
}

\newcommand\orcidicon[1]{\href{https://orcid.org/#1}{\mbox{\scalerel*{
\begin{tikzpicture}[yscale=-1,transform shape]
\pic{orcidlogo};
\end{tikzpicture}
}{|}}}}

\tikzset{
  block/.style={rectangle, draw, fill=blue!10, text centered,
                rounded corners=2pt, minimum height=1.8em, font=\scriptsize,
                text width=4.0em},
  blockw/.style={rectangle, draw, fill=orange!15, text centered,
                 rounded corners=2pt, minimum height=1.8em, font=\scriptsize,
                 text width=4.0em},
  blockf/.style={rectangle, draw, fill=green!12, text centered,
                 rounded corners=2pt, minimum height=1.8em, font=\scriptsize,
                 text width=4.2em},
  blockv/.style={rectangle, draw, fill=gray!12, text centered,
                 rounded corners=2pt, minimum height=1.8em, font=\scriptsize,
                 text width=4.0em},
  arr/.style={-{Stealth[length=4pt]}, thick},
  skipconn/.style={dashed, -{Stealth[length=4pt]}, thick, gray},
  lossbox/.style={rectangle, draw=red!60, fill=red!6, dashed,
                  rounded corners=2pt, minimum height=1.8em, font=\scriptsize,
                  text width=4.8em},
}

\title{CLEAR-IR: Clarity-Enhanced Active Reconstruction of Infrared Imagery for Low-light Enhancement}

\author{Nathan Shankar$^{\orcidicon{0000-0002-7255-2985}}$, Marton Gonczy$^{\orcidicon{0009-0001-0319-5400}}$, Hujun Yin$^{\orcidicon{0000-0002-9198-5401}}$, and Pawel Ladosz$^{\orcidicon{0000-0002-1154-8333}}$%
\thanks{This work was supported by Robotics and Artificial Intelligence Collaboration (RAICo).
(Corresponding author: Pawel Ladosz.)
The authors are with the the Mechanical, Aerospace and Civil Engineering and the Electrical and Electronics Departments at the University of Manchester, Manchester, United Kingdom, M13 9PL. (e-mail: pawel.ladosz@manchester.ac.uk).
}}

\begin{document}

\maketitle
\thispagestyle{empty}
\pagestyle{empty}

\begin{abstract}
This paper presents CLEAR-IR (Clarity Enhanced Active Reconstruction of Infrared Imagery), a novel framework for enabling robust robotic perception in dark environments using active infrared~(IR) sensing. IR streams are less susceptible to noise than RGB in low light conditions but are dominated by structured light emitter patterns that hinder high level tasks such as object detection, tracking, and localisation. CLEAR-IR addresses this by reconstructing clean, RGB styled images from emitter populated IR inputs using a dual stream architecture inspired by DeepMAO~\cite{sikdar2023deepmao}, enabling standard RGB trained vision pipelines to operate without visible light. Evaluation across object detection, ArUco marker detection, and visual SLAM demonstrates that CLEAR-IR produces reliable perception in conditions ranging from well lit to completely dark, where passive RGB based enhancement methods fail entirely due to the physical absence of sensor signal.
\end{abstract}

\begin{keywords}
Infrared Imaging, Image Reconstruction, U-Net, Low Light Perception, Computer Vision, Robotics
\end{keywords}

\section{Introduction}
Lighting invariant vision systems are desirable for enabling robots to operate robustly across diverse and unpredictable environments without requiring modifications to the underlying perception pipeline. In order to support high level tasks such as object detection, semantic segmentation, and image classification, the vision system must remain reliable even in low light or completely dark scenes. Such capabilities are critical in domains like mine shaft exploration, post disaster victim identification, nuclear facility inspection, and visual loop closure in feature deprived environments using aruco markers.

Conventional RGB cameras, while effective under well lit conditions, suffer from significant degradation in low light scenes~\cite{guo2023low}. Image quality is compromised due to sensor level limitations, such as dark current and readout noise. Attempts to compensate using an increased exposure time introduce motion blur, rendering them ineffective for dynamic environments. Furthermore, data driven approaches relying solely on visible spectrum low light and long exposure pairs are fundamentally limited by information theoretic constraints. If critical scene information is not captured by the sensor due to extreme darkness or sensor level noise, it is challenging to reconstruct accurately. Consequently, the efficacy of these networks remains strictly bound by the inherent observability of the visible spectrum in degraded conditions.

Another approach involves equipping robots with onboard illumination systems~\cite{crocetti2025active}. However, in environments such as tunnels or disaster zones, the Tyndall effect scatters visible light off airborne particles, saturating the camera with artefacts and obscuring key features. The approach can also suffer from overexposure, specular reflections, and higher power usage. In addition, active illumination can create strong shadows that hide structural details and hinder tasks such as object detection and localisation.

\begin{figure}[htbp]
    \centering
  \subfloat[Low light room captured using an RGB camera\label{1a}]{%
       \includegraphics[width=0.48\linewidth]{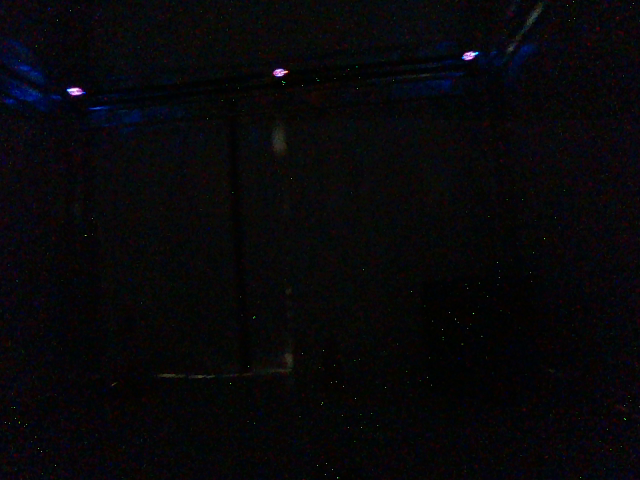}}
    \hfill
  \subfloat[Low light room captured using an IR camera\label{1b}]{%
        \includegraphics[width=0.48\linewidth]{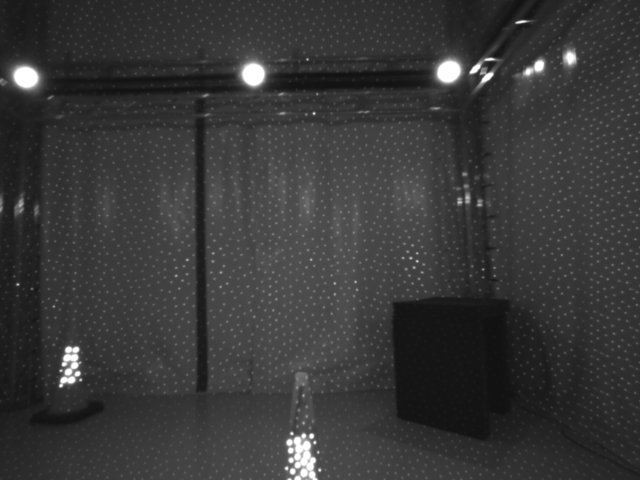}}
  \caption{The same low light scene captured with RGB (left) and active IR (right). The RGB image contains no usable visual information, whereas the IR image clearly reveals scene structure despite the absence of ambient light.}
  \label{fig:dark_scene}
\end{figure}

To address these challenges, this work explores the use of active infrared~(IR) imaging, which operates independently of ambient illumination. As seen in \autoref{fig:dark_scene}, the dark scene captured by the RGB camera fails to present good visual cues, while the same scene in the IR spectrum reveals several objects. Although active IR systems penetrate some scattering environments better than visible light sources because of their longer wavelengths, the controlled laboratory experiments in this work focus on low light and completely dark indoor conditions rather than particulate conditions and validating performance in dust or smoke is identified as an important avenue for future work. Passive IR sensors avoid structured light artefacts but are expensive and usually offer low resolution. Active IR systems, meanwhile, often use structured light or flash patterns that can interfere with feature extraction and disrupt tasks such as object detection and SLAM.

\begin{figure*}[b!]
    \centering
    \subfloat[Emitter laser power sweep \label{fig:em_power}]{%
    \includegraphics[width=1.5\columnwidth]{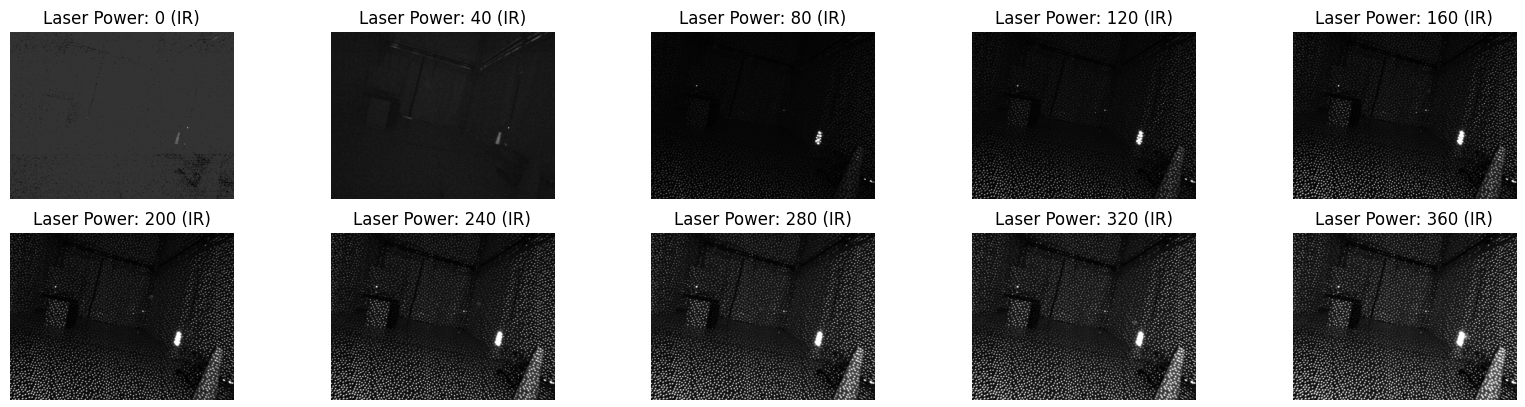}}

    \caption{Increasing laser power progresses from invisible dots to a fully populated pattern.}
    \label{fig:sota_comparison}
\end{figure*}
To overcome this, CLEAR-IR is proposed. The framework reconstructs clean and interpretable IR images from raw emitter populated inputs using a dual stream architecture~\cite{sikdar2023deepmao} that suppresses structured light noise while preserving spatial and textural details critical for vision based tasks. The output is a visually coherent IR image that serves as a robust input to standard RGB trained perception modules, enabling their use without any visible light illumination. The contributions of this work are:
\begin{itemize}
    \item CLEAR-IR, a novel dual stream architecture tailored for reconstructing clean IR images by suppressing structured light artefacts, is introduced.
    \item A composite loss function specifically selected to accommodate the spectral disparity and lack of pixel perfect alignment between IR and RGB inputs, enabling robust pattern removal while preserving details critical for robotic perception.
    \item Evaluation of IR image reconstruction for robotics is presented, demonstrating significant improvements across key tasks such as object detection and marker detection.
    \item Comparison with other SOTA techniques for localisation in extreme low light scenes.
\end{itemize}

\section{Related Work}
Vision based robotic perception in challenging lighting conditions has been an active area of research. One line of work has focused on improving the performance of standard RGB cameras in low light. This includes sensor level enhancements such as improved read out noise reduction and pixel architectures~\cite{ma2022review}, as well as computational photography techniques such as multiexposure fusion~\cite{ying2017bio} and burst photography~\cite{karadeniz2021burst}. Although these techniques can improve image quality, they are often computationally intensive and can introduce motion blur or require static scenes, limiting their use in dynamic robotic applications.

A separate body of work leverages deep learning for low light image enhancement. Prominent examples include supervised learning methods that train neural networks on paired low light and well lit images~\cite{chobola2024fast,reis2025low}. However, these approaches are highly dependent on the quality and diversity of the training data and may not generalise well to different camera hardware or unseen environments. Unsupervised and semi supervised methods have also been explored~\cite{yang2020fidelity,fahim2023semi,kandula2023unsupervised}, but they often struggle to remove complex artefacts such as structured light patterns found in active IR systems. Generative adversarial networks~(GANs) have been used for low light enhancement~\cite{fu2022gan,wang2024low,he2025swinlightgan}, but these methods do not translate well to extremely dark scenes, which require some form of active sensing. Moreover, artificial artefacts introduced by these methods can cause SLAM pipelines to latch onto phantom features, leading to tracking drift and map corruption.

Active infrared systems provide a robust alternative for dark scene perception by illuminating the environment with a dedicated source. While high power IR floodlights offer uniform coverage, they frequently cause severe near field saturation and shadows that occlude critical features. To mitigate this, depth camera manufacturers utilise structured light or modulated laser patterns~\cite{Alhwarin_Ferrein_Scholl_2014} that distribute energy into discrete, evenly spaced points, maintaining a higher usable dynamic range even as pattern density increases at close proximity. Note that these emitter patterns also obey the inverse square law but the advantage over floodlights is not a power law benefit but rather that concentrated spots avoid the global image plane saturation characteristic of flood illumination.
These active IR systems are already standard components in many robotic platforms, so leveraging them for 2D perception minimises system complexity, payload weight, and maintenance costs. However, these high frequency patterns present a significant challenge for 2D vision tasks, as standard algorithms often misinterpret them as scene texture, leading to false detections. Previous denoising works~\cite{guan2020fixed,lee2023infrared,barral2024fixed} focus primarily on removing static sensor level artefacts rather than the dynamic, pattern dependent noise inherent to active emitters.

In contrast to existing methods, this work addresses the specific problem of structured light artefacts in active IR imagery, proposing a dual stream deep learning architecture that learns to suppress these patterns while preserving the semantic information needed for high level vision tasks.

\section{Methodology}
This section presents the CLEAR-IR framework. Unlike standard encoder decoder approaches, CLEAR-IR uses a hybrid dual branch architecture to effectively remove structured light artefacts while preserving underlying visual details.

\subsection{Emitter Pattern Characteristics}\label{sec:em_characteristics}
Active infrared (IR) systems enable vision in dark environments by projecting a pseudo random, semi deterministic dot pattern onto the scene. The raw IR image is therefore populated with these dots, which standard vision algorithms misidentify as scene texture, complicating object detection and segmentation.

\begin{figure}[htbp]
\centering
\subfloat[Reflectivity \label{fig:reflectivity}]{%
    \includegraphics[width=0.45\columnwidth]{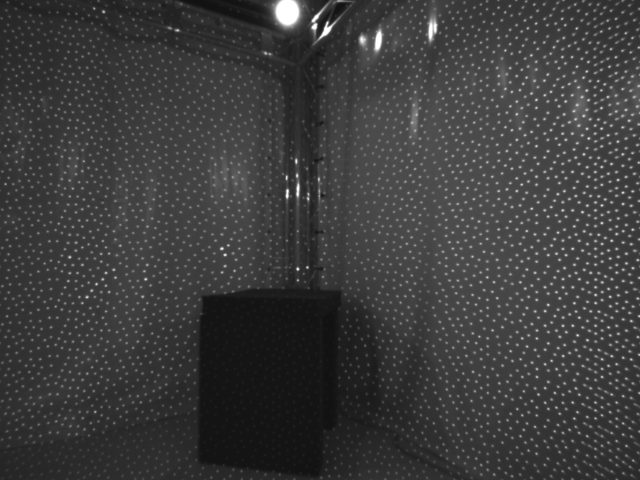}}
\hfill
\subfloat[Distance \label{fig:distance}]{%
    \includegraphics[width=0.45\columnwidth]{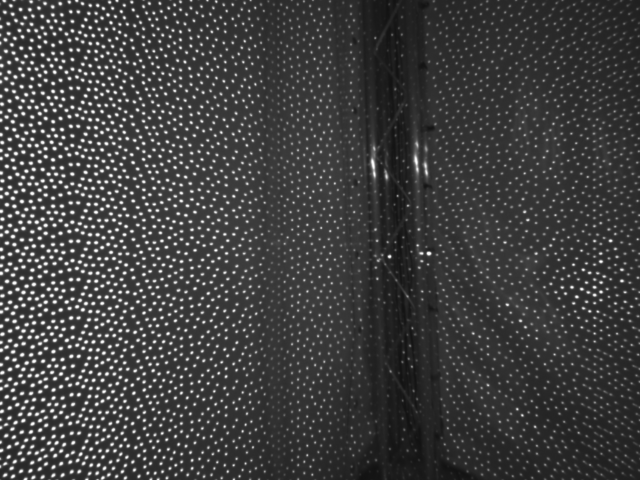}}
\\
\subfloat[Bright room \label{fig:bright_ir}]{%
    \includegraphics[width=0.45\columnwidth]{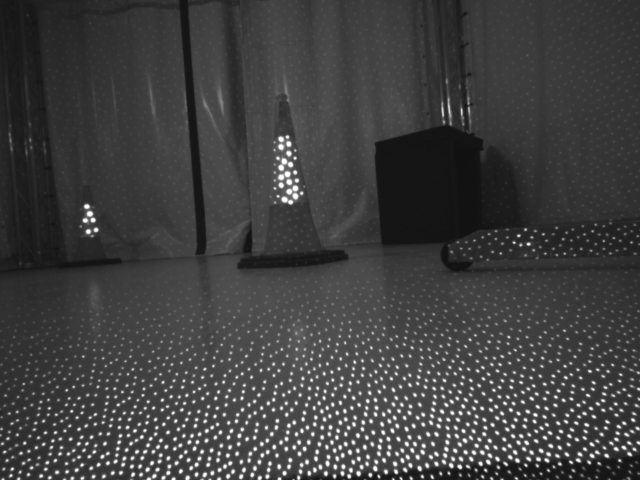}}
\hfill
\subfloat[Dark room \label{fig:dark_ir}]{%
    \includegraphics[width=0.45\columnwidth]{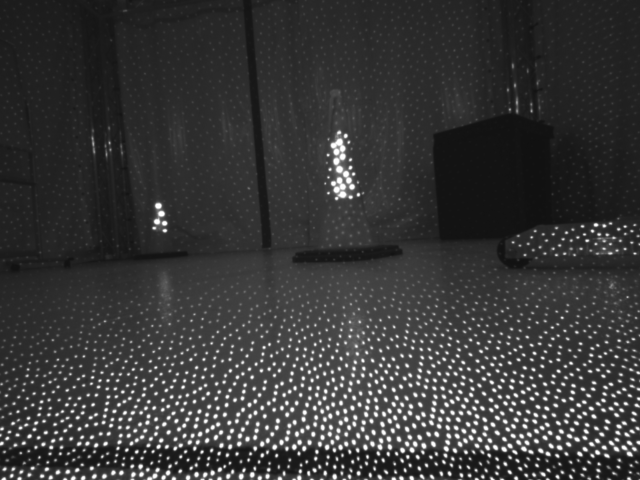}}
\caption{Key factors influencing emitter pattern appearance. (a) High reflectivity surfaces (plastic) show denser patterns than low reflectivity ones (foam). (b) Dots are larger and more intense at close range. (c,d) Ambient light suppresses effective emitter visibility and the pattern is much stronger in darkness.}
\label{fig:five_subfigs}
\end{figure}

The intensity of the emitter dots is non uniform and depends on four main factors:
\begin{itemize}
\item \textbf{Emitter power:} A higher laser power yields a denser, more defined pattern (\autoref{fig:em_power}).
\item \textbf{Object reflection:} Objects with higher IR reflectivity appear brighter, reflecting a sharper pattern (\autoref{fig:reflectivity}).
\item \textbf{Distance to the emitter:} Dot intensity decreases with the square of distance as the dots appear larger and denser at close range (\autoref{fig:distance}).
\item \textbf{Room brightness:} Ambient light dims the effective emitter visibility and the same pattern appears sharper in darker scenes (\autoref{fig:bright_ir}, \autoref{fig:dark_ir}).
\end{itemize}

This variability across scenes motivates the use of a learned, perceptual loss rather than simple pixel wise reconstruction.

\subsection{CLEAR-IR Architecture}

CLEAR-IR uses a dual stream architecture that processes the IR input in parallel through two complementary branches which includes a Context Stream and a Detail Stream. The architecture is depicted in \autoref{fig:architecture}.

\begin{figure*}[htbp]
\centering
\resizebox{\textwidth}{!}{%
\begin{tikzpicture}[node distance=0.6cm and 0.8cm]

  \node[blockv] (input) {IR Input\\$640{\times}480{\times}1$};

  \node[block, right=1.0cm of input, yshift=2.3cm] (enc1) {Double Conv\\32 ch.};
  \node[block, right=0.55cm of enc1] (enc2) {Double Conv\\64 ch.};
  \node[block, right=0.55cm of enc2] (enc3) {Double Conv\\128 ch.};
  \node[block, right=0.55cm of enc3] (bot) {Bottleneck\\256 ch. + Drop};
  \node[block, right=0.55cm of bot] (dec3) {Double Conv\\128 ch.};
  \node[block, right=0.55cm of dec3] (dec2) {Double Conv\\64 ch.};
  \node[block, right=0.55cm of dec2] (dec1) {Conv Block\\32 ch.};

  \draw[skipconn] (enc1.south) -- ++(0,-0.60) -| (dec1.south) node[midway, below, font=\tiny] {};
  \draw[skipconn] (enc2.south) -- ++(0,-0.42) -| (dec2.south);
  \draw[skipconn] (enc3.south) -- ++(0,-0.25) -| (dec3.south);

  \draw[arr] (input.north) |- (enc1.west);
  \draw[arr] (enc1) -- node[above,font=\tiny]{} (enc2);
  \draw[arr] (enc2) -- node[above,font=\tiny]{} (enc3);
  \draw[arr] (enc3) -- node[above,font=\tiny]{} (bot);
  \draw[arr] (bot) -- node[above,font=\tiny]{} (dec3);
  \draw[arr] (dec3) -- node[above,font=\tiny]{} (dec2);
  \draw[arr] (dec2) -- node[above,font=\tiny]{} (dec1);

  \node[blockw, right=1.0cm of input, yshift=-2.3cm] (oc1) {Conv 3$\times$3\\32 ch.};
  \node[blockw, right=0.55cm of oc1] (oc2) {Conv 3$\times$3\\32 ch.};
  \node[blockw, right=0.55cm of oc2] (oc3) {Conv 3$\times$3\\32 ch.};

  \draw[arr] (enc1.south) -- ++(0,-0.95) -| (oc1.north) node[pos=0.15, left, font=\tiny] {};
  \draw[arr] (oc1) -- (oc2);
  \draw[arr] (oc2) -- (oc3);

  \node[font=\small\itshape\bfseries, above=0.15cm of bot] {Context Stream};
  \node[font=\small\itshape\bfseries, below=0.15cm of oc2] {Detail Stream};

  \node[blockf, right=2.0cm of dec1, yshift=-2.3cm] (fuse) {Add Fusion\\($\oplus$)};
  \node[blockf, right=0.55cm of fuse] (refine) {Conv\\32 ch.};
  \node[blockv, right=0.55cm of refine] (output) {Recon. Output\\$640{\times}480{\times}1$};

  \draw[arr] (dec1.east) -- ++(0.3,0) |- (fuse.north);
  \draw[arr] (oc3.east)  -- ++(0.3,0) |- (fuse.south);
  \draw[arr] (fuse)    -- (refine);
  \draw[arr] (refine)  -- (output);

  \node[lossbox, below=2.2cm of refine] (vgg) {$\epsilon\mathcal{L}_{\mathrm{perceptual}}$};
  \node[lossbox, left=0.8cm of vgg] (reg) {$\zeta\mathcal{L}_{\mathrm{tv}}$};
  \node[lossbox, left=0.8cm of reg] (spatial) {$\alpha\mathcal{L}_{\mathrm{mae}} + \beta\mathcal{L}_{\mathrm{ssim}}$};
  \node[lossbox, right=0.8cm of vgg] (detail) {$\gamma\mathcal{L}_{\mathrm{freq}} + \delta\mathcal{L}_{\mathrm{sobel}}$};
  
  \node[lossbox, right=2.2cm of detail] (loss) {$\mathcal{L}_\mathrm{total}$};

  \draw[dashed, -{Stealth[length=3pt]}, red!60] (output.south) -- ++(0,-0.8) -| (spatial.north) node[pos=0.2, above, font=\tiny] {};
  \draw[dashed, -{Stealth[length=3pt]}, red!60] (output.south) -- ++(0,-0.8) -| (detail.north) node[pos=0.2, above, font=\tiny] {};
  \draw[dashed, -{Stealth[length=3pt]}, red!60] (output.south) -- ++(0,-0.8) -| (reg.north) node[pos=0.2, above, font=\tiny] {};
  \draw[dashed, -{Stealth[length=3pt]}, red!60] (output.south) -- ++(0,-0.8) -| ([xshift=-0.4cm]vgg.north) node[pos=0.15, above, font=\tiny] {};
  
  \draw[dashed, -{Stealth[length=3pt]}, red!60] (vgg.south) -- ++(0,-0.4) -| (loss.south) node[pos=0.2, below, font=\tiny] {Perceptual Quality};
  \draw[dashed, -{Stealth[length=3pt]}, red!60] (spatial.south) -- ++(0,-1.2) -| (loss.south) node[pos=0.2, below, font=\tiny] {Spatial Fidelity};
  \draw[dashed, -{Stealth[length=3pt]}, red!60] (reg.south) -- ++(0,-0.8) -| (loss.south) node[pos=0.2, below, font=\tiny] {Regularisation};
  \draw[dashed, -{Stealth[length=3pt]}, red!60] (detail.east) -- (loss.west) node[pos=0.5, below, font=\tiny] {Detail Preservation};

\end{tikzpicture}%
}
\caption{CLEAR-IR Architecture}
\label{fig:architecture}
\end{figure*}

The architecture adapts the dual stream principle of DeepMAO~\cite{sikdar2023deepmao}, which was originally designed for satellite building segmentation with a pretrained EfficientNet-B3 encoder and classification losses. Three key modifications are made to suit IR image reconstruction are as follows:
\begin{itemize}
    \item The EfficientNet-B3 encoder is replaced with a lightweight standard U-Net \cite{ronneberger2015u} encoder using convolutional blocks with increasing filter depths.
    \item A multi scale overcomplete (MSO) module, which in the original uses non integer dilated interpolations, is replaced with a cascade of full resolution $3{\times}3$ convolutions without any striding or pooling
    \item The training objective is changed from pixel level classification to intensity regression using a combined perceptual and structural loss.
\end{itemize}

As shown in the ablation study (\autoref{fig:detections}), deploying the original architecture for IR reconstruction causes over smoothing and loss of edge fidelity.

\subsubsection{Context Stream (U-Net Backbone)}
The first branch captures global context and low frequency structural information.
\begin{itemize}
    \item \textbf{Hierarchical encoding:} Convolutional blocks with increasing filter depths and max pooling expand the receptive field to capture global scene geometry.
    \item \textbf{Regularisation:} Dropout in deeper encoder layers and the bottleneck prevents overfitting to specific emitter characteris.
    \item \textbf{Decoder and skip connections:} Transposed convolutions upsample feature maps, concatenated with corresponding encoder features via skip connections to recover spatial detail lost during pooling.
\end{itemize}

\subsubsection{Detail Stream (Overcomplete Branch)}
While pooling in the U-Net suppresses emitter patterns, it also destroys true high frequency scene detail. The second branch operates as a full resolution overcomplete pathway.
\begin{itemize}
    \item \textbf{Full spatial resolution:} Starting from the initial 32 filter feature maps, a cascade of $3{\times}3$ convolutional layers processes the input without any scaling, pooling, or strided operations, maintaining a strict 1:1 input to output spatial correspondence.
    \item \textbf{Detail preservation:} By bypassing downsampling, this branch isolates genuine high frequency structural boundaries from the emitter noise, injecting fine scene textures into the final reconstruction.
\end{itemize}

\subsubsection{Feature Fusion}
The outputs of both branches are combined using additive residual fusion:
\begin{equation}
    F_\text{fused} = F_\text{unet} \oplus F_\text{overcomplete}
\end{equation}
where $\oplus$ denotes element wise addition. The fused features are passed through a final convolutional refinement block to generate the reconstructed IR image.

\subsection{Loss Function}
Despite extrinsic calibration and warping, minor alignment residuals persist due to parallax and the sensor baseline between the IR and RGB cameras. Therefore, a composite loss is used that prioritises perceptual quality and structural integrity over simple pixel wise accuracy:
\begin{gather}
\label{total_loss}
\begin{multlined}
\mathcal{L}_\text{total} = \alpha\mathcal{L}_{mae} + \beta\mathcal{L}_{ssim} + \gamma\mathcal{L}_{freq} \\
+ \delta\mathcal{L}_{sobel} + \epsilon\mathcal{L}_{perceptual} + \zeta\mathcal{L}_{tv}
\end{multlined}
\end{gather}
The components are:
\begin{itemize}
    \item \textbf{Spatial fidelity ($\mathcal{L}_{mae}$, $\mathcal{L}_{ssim}$):} Mean Absolute Error for base intensity matching and SSIM~\cite{Wang_2004} to preserve luminance and contrast structure.
    \item \textbf{Detail preservation ($\mathcal{L}_{freq}$, $\mathcal{L}_{sobel}$):} $\mathcal{L}_{freq}$ minimises the $L_1$ distance between Laplacian filtered prediction and target while $\mathcal{L}_{sobel}$ penalises the gradient discrepancies.
    \item \textbf{Perceptual quality ($\mathcal{L}_{perceptual}$):} A VGG19 network pre trained on ImageNet~\cite{Imagenet_2009} is used as a frozen feature extractor during training only and it is not part of the inference graph. Intermediate activations are extracted from both the predicted output and the aligned ground truth RGB image, and the $L_1$ distance between these feature maps is minimised, ensuring the reconstruction aligns with perceptually meaningful structure~\cite{Johnson_2016}.
    \item \textbf{Regularisation ($\mathcal{L}_{tv}$):} Total Variation loss promotes spatial smoothness in uniform regions.
\end{itemize}

The weighting coefficients ($\alpha, \beta, \gamma, \delta, \epsilon, \zeta$) were experimentally tuned to prioritise $\mathcal{L}_{ssim}$ and $\mathcal{L}_{perceptual}$, ensuring the model focuses on structural removal of the emitter pattern rather than overfitting to minor pixel misalignments. 

\subsection{Data Augmentation and Training Implementation}
A custom dataset was developed by capturing 6,719 paired active IR and greyscale RGB images using an Intel RealSense D455 camera~\cite{Intel_2020}. All data augmentation was applied to the original 6,719 base pairs before the train/test split, so no augmented images appear in the test set. Factory calibrated extrinsics were used to align the two sensors. Augmentation was done using random rotations $\pm30^\circ$, flipping, affine transforms and brightness adjustments to expanded the base set to 33,595 images, split 90:10 for training and evaluation. The model was trained to map IR inputs to an RGB style ground truth, enabling downstream vision tasks originally developed for the visible spectrum to run on reconstructed IR images. Training used the Adam optimiser with a cosine decay scheduler at $640{\times}480$ resolution. Early stopping was also used to ensure convergence while preventing overfitting.

\section{Experimental Results}

\subsection{Evaluation Framing}
\label{sec:eval_framing}
A key design decision in this evaluation is how to compare CLEAR-IR against RGB based enhancement methods. These methods operate on passively captured, nearly dark RGB streams. In extreme low light, the failure of such methods is a physical consequence of the sensor receiving a noisy signal, not an algorithmic weakness. Comparing CLEAR-IR against these methods in darkness therefore demonstrates the limits of passive sensing rather than proving algorithmic superiority. Accordingly, the RGB methods are presented as a context baseline, and the primary algorithmic comparisons are:
\begin{itemize}
    \item CLEAR-IR versus Raw IR which consists of information from the same sensor.
    \item CLEAR-IR versus the U-Net baseline which has the same modality, same active IR input but different architecture
\end{itemize}
In low light conditions where some ambient illumination is still present, both classes of method receive usable signal, enabling a fairer cross modal comparison. The CLEAR-IR architecture was trained from scratch on the custom IR to RGB dataset. The RGB based SOTA methods (Zero-DCE, LLFormer, CIDNet, Retinex, CLAHE) had open sourced models and weights and retraining them on IR data was not performed since they are fundamentally designed for the visible spectrum. The DeepMAO ablation baseline was retrained from scratch on the same dataset with its original EfficientNet-B3 encoder retained, so that only the MSO module and training objective differ from CLEAR-IR.

This section presents a comprehensive evaluation focusing on the effectiveness of CLEAR-IR in preparing infrared images for high level computer vision tasks.

The advantages of CLEAR-IR are emphasised through a quantitative comparison against traditional baselines and state of the art (SOTA) enhancement methods, including CLAHE~\cite{CLAHE_Srikanth}, Retinex inspired methods~\cite{Cai_2017_ICCV}, Zero-DCE~\cite{Zero-DCE}, LLFormer~\cite{wang2023ultra}, and CIDNet~\cite{yan2025hvi}.

\begin{figure*}[htbp]
    \centering
    \subfloat[Ground Truth RGB\label{fig:gt_rgb}]{%
        \includegraphics[width=0.18\linewidth]{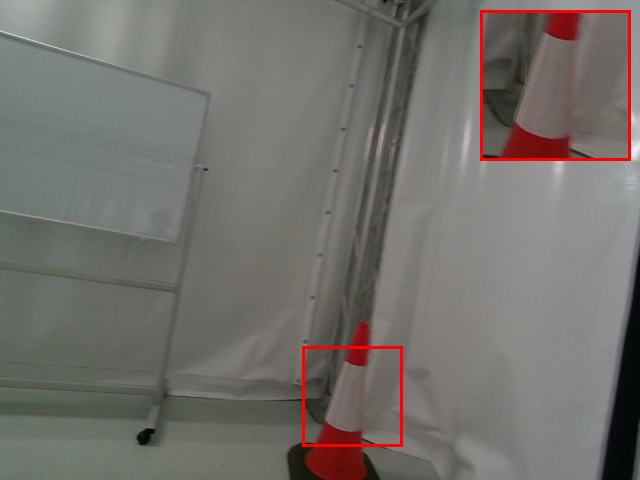}}
    \hfill
    \subfloat[Dark RGB\label{fig:rgb}]{%
        \includegraphics[width=0.18\linewidth]{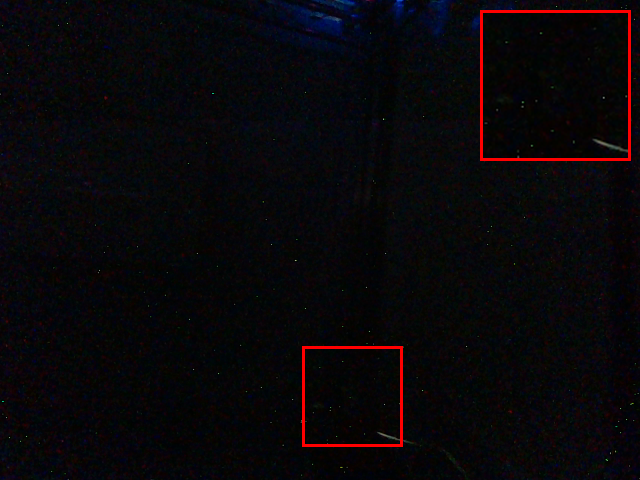}}
    \hfill
    \subfloat[CLAHE\label{fig:clahe_out}]{%
        \includegraphics[width=0.18\linewidth]{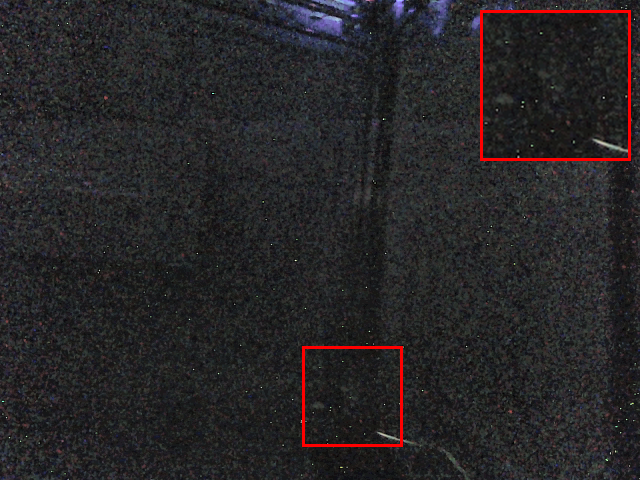}}
    \hfill
    \subfloat[Retinex\label{fig:rettheo}]{%
        \includegraphics[width=0.18\linewidth]{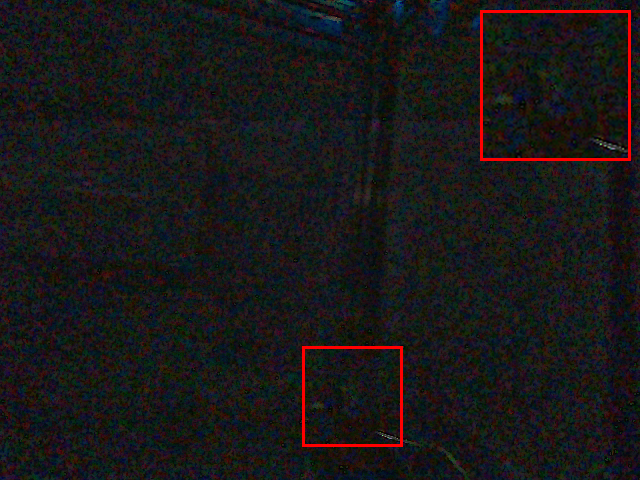}}
    \hfill
    \subfloat[Zero-DCE\label{fig:zerodce_out}]{%
        \includegraphics[width=0.18\linewidth]{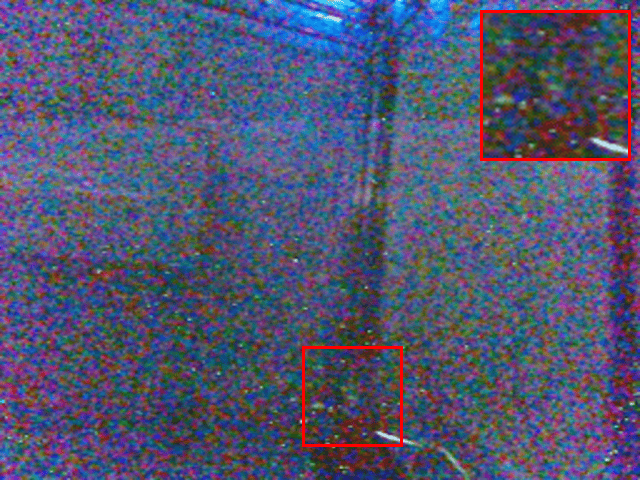}}

    \subfloat[LLFormer\label{fig:llformer_out}]{%
        \includegraphics[width=0.18\linewidth]{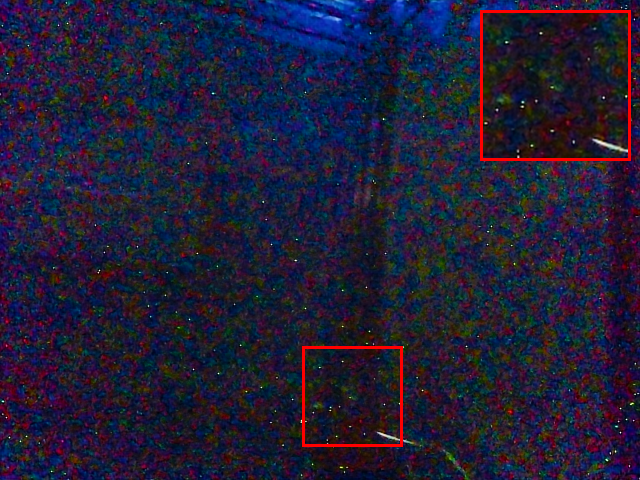}}
    \hfill
    \subfloat[CIDNet\label{fig:cidnet_out}]{%
        \includegraphics[width=0.18\linewidth]{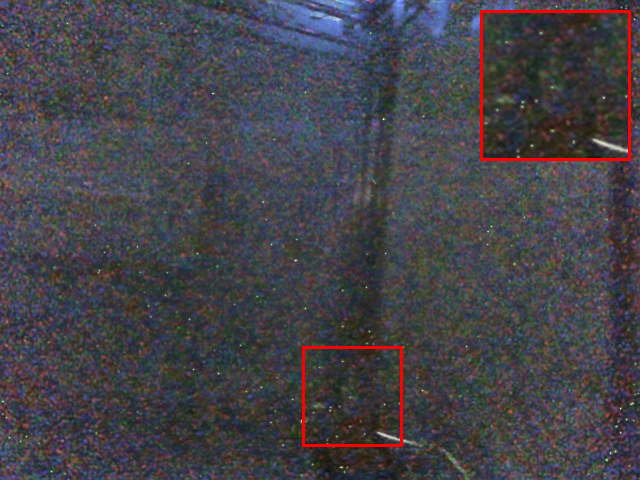}}
    \hfill
    \subfloat[Raw IR\label{fig:raw_ir_out}]{%
        \includegraphics[width=0.18\linewidth]{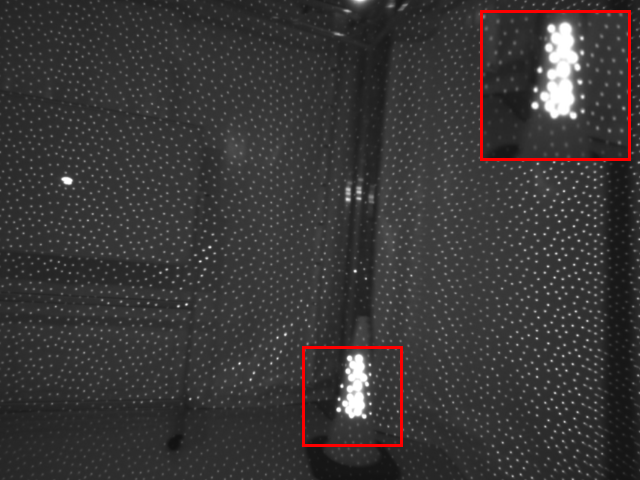}}
    \hfill
    \subfloat[U-Net\label{fig:unet_out}]{%
        \includegraphics[width=0.18\linewidth]{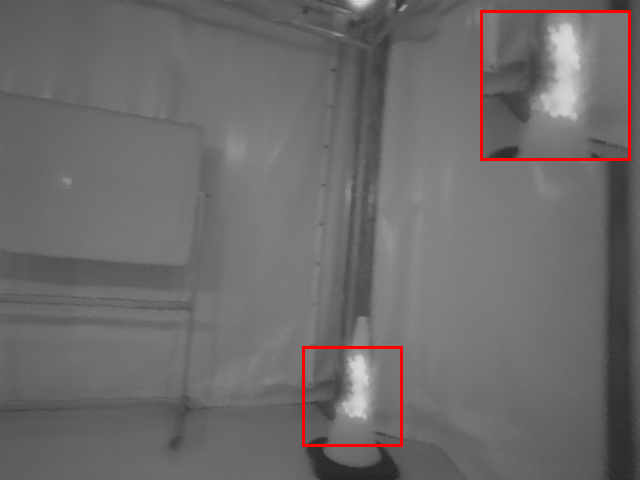}}
    \hfill
    \subfloat[CLEAR-IR (Ours)\label{fig:our_denoised_out}]{%
        \includegraphics[width=0.18\linewidth]{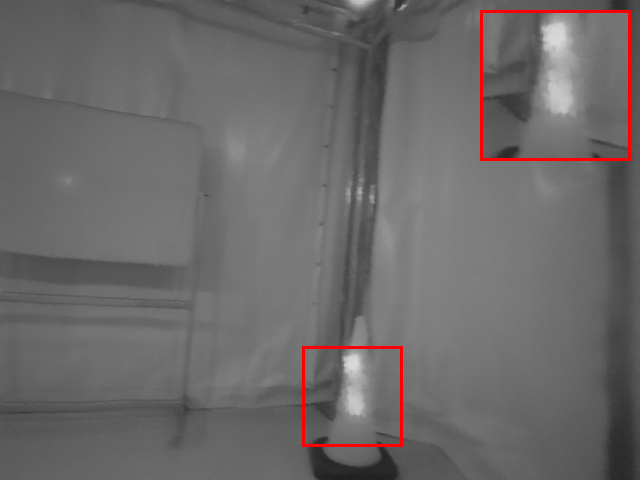}}

    \caption{Visual comparison of enhancement methods on an unseen scene. RGB based methods (c--g) are applied to the dark passive RGB stream and exhibit noise and colour artefacts. Raw IR (h) retains emitter dot patterns throughout. The U-Net baseline (i) removes most patterns but produces some residual blurring in reflective regions. CLEAR-IR (j) produces a clean, structurally coherent reconstruction without introducing hallucinated features. See zoomed crops in supplementary material for edge region detail.}
    \label{fig:sota_comparison}
\end{figure*}

The visual results in \autoref{fig:sota_comparison} show the outputs from different techniques on an unseen scene. While other methods attempt to enhance the image, the enhanced RGB images are affected by noise inherent in low light scenes. The CLEAR-IR framework produces a clean, artefact free image that maintains the scene structural integrity without introducing spurious features or hallucinating content.

\begin{table*}[b!]
\centering
\caption{Performance comparison of image enhancement methods on high level robotics perception tasks}
\label{tab:cv_performance}
\begin{tabular}{lccccc}
\toprule
\textbf{Method}
& \multicolumn{3}{c}{\textbf{Object Detection (Visible Objects)}}
& \multicolumn{2}{c}{\textbf{ArUco Marker Detection}} \\
\cmidrule(lr){2-4} \cmidrule(lr){5-6}
& Detected & Missed & Mislabelled & Correct & Missed \\
\midrule
\multicolumn{6}{l}{\textit{RGB based methods (passive dark RGB stream)}} \\
Retinex Theory~\cite{Cai_2017_ICCV}     & 1  & 9  & 0 & 5           & 1           \\
CLAHE~\cite{CLAHE_Srikanth}             & 1  & 9  & 0 & 3           & 3           \\
Zero-DCE~\cite{Zero-DCE}               & 0  & 10 & 0 & 5           & 1           \\
LLFormer~\cite{wang2023ultra}           & 0  & 10 & 0 & 5           & 1           \\
CIDNet~\cite{yan2025hvi}               & 1  &  9 & 0 & \textbf{6}  & \textbf{0}  \\
\midrule
\multicolumn{6}{l}{\textit{IR based methods (active IR stream)}} \\
Raw IR                                  & 0  & 9  & 1 & 0           & 6           \\
DeepMAO (ablation)                     & 2  & 8  & 2 & 0           & 6           \\
U-Net (ablation)                       & 7  & 3  & 0 & \textbf{6} & \textbf{0} \\
CLEAR-IR (Proposed)                    & \textbf{11} & \textbf{0}  & 2          & \textbf{6} & \textbf{0} \\
\bottomrule
\end{tabular}
\end{table*}

\autoref{tab:cv_performance} presents a quantitative comparison on high level perception tasks. A critical distinction is drawn between methods that simply improve image aesthetics and those that produce a representation suitable for downstream algorithms. The proposed work is also used in feature tracking and visual odometry to highlight stability across frames in unseen environments.

\subsection{Ablation Study}
\autoref{tab:cv_performance} and \autoref{fig:detections} compare CLEAR-IR against Raw IR, the original DeepMAO architecture retrained on the IR dataset, and a standalone U-Net baseline. The original DeepMAO reconstruction is blurry and looses edge fidelity, causing both ArUco and object detection to fail. The U-Net baseline improves substantially but produces residual jitter on highly reflective surfaces such as traffic cones. CLEAR-IR, which adds the overcomplete detail stream to the U-Net backbone, recovers 11 out of 11 objects and all 6 markers. Since all other training details are held constant, the gain is attributable to the overcomplete branch.

\subsection{Object Detection and Segmentation}
To demonstrate the utility of the reconstructed images for advanced perception tasks, a pre trained YOLOv26 object detection and segmentation model was used for evaluation. This model was not trained on IR data, making this a zero shot evaluation to determine if the proposed preprocessing step can make the images usable with standard RGB based vision pipelines.

\begin{figure}[htbp]
    \centering
    \subfloat[YOLOv26 on RGB image]{%
        \includegraphics[width=0.45\columnwidth]{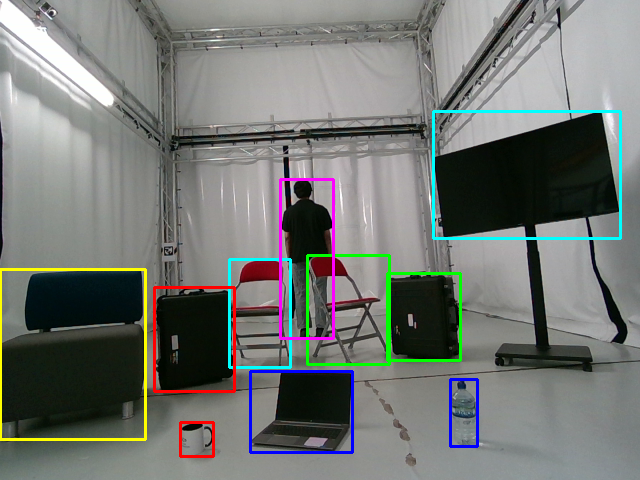}}
    \hfill
    \subfloat[YOLOv26 on raw IR image]{%
        \includegraphics[width=0.45\columnwidth]{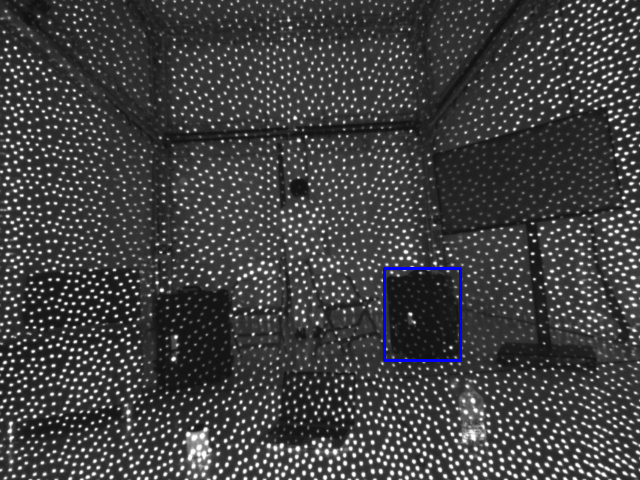}}
    \hfill
    \subfloat[YOLOv26 on U-Net image]{%
        \includegraphics[width=0.45\columnwidth]{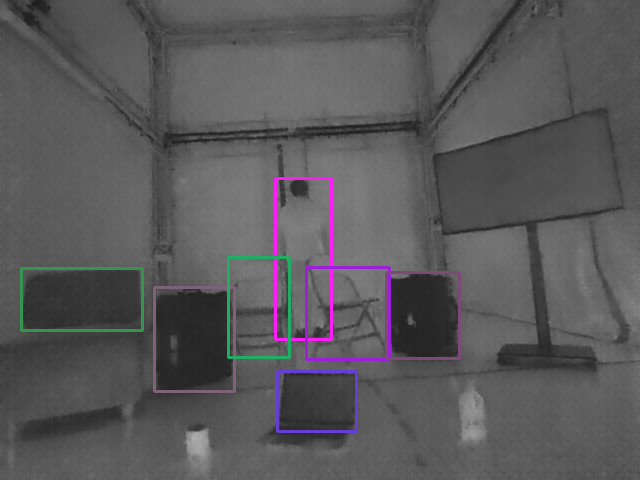}}
    \hfill
    \subfloat[YOLOv26 on CLEAR-IR image]{%
        \includegraphics[width=0.45\columnwidth]{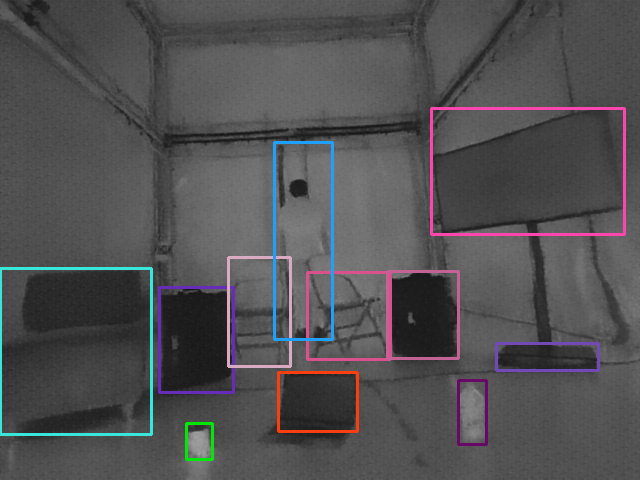}}
    \caption{Object detection results. YOLOv26 fails entirely on raw IR (b) due to emitter dot patterns. On U-Net output (c), most objects are detected but three are missed. On CLEAR-IR output (d), all 11 objects are detected but some mislabelling occurs due to the greyscale nature of IR, but detections are consistent across frames.}
    \label{fig:yolo_results}
\end{figure}

On the raw IR images, YOLOv26 failed to detect or segment objects. The emitter patterns and resulting image noise prevented the model from recognising the distinct shapes and features of common objects. After preprocessing the images with the CLEAR-IR model, the performance of YOLOv26 dramatically improved. As illustrated in \autoref{fig:yolo_results}, the model successfully detected and segmented objects in the reconstructed images, which were completely missed in the raw inputs. While tested on a variety of images, some labels were incorrect since the YOLOv26 model had not been trained on the specific objects present in the scene. For example, traffic cones were occasionally misclassified as bottles, and a white storage cabinet was misclassified as a refrigerator. Although the greyscale nature of IR data makes some misclassification inevitable without fine tuning. Importantly, despite some label mismatches, the model was able to consistently detect and track these objects across frames, demonstrating the usability of CLEAR-IR outputs for object level perception.

\subsection{ArUco Marker Detection}
\begin{figure}[htbp]
    \centering
    \subfloat[RGB image (reference)\label{fig:detections:rgb}]{%
        \includegraphics[width=0.45\linewidth]{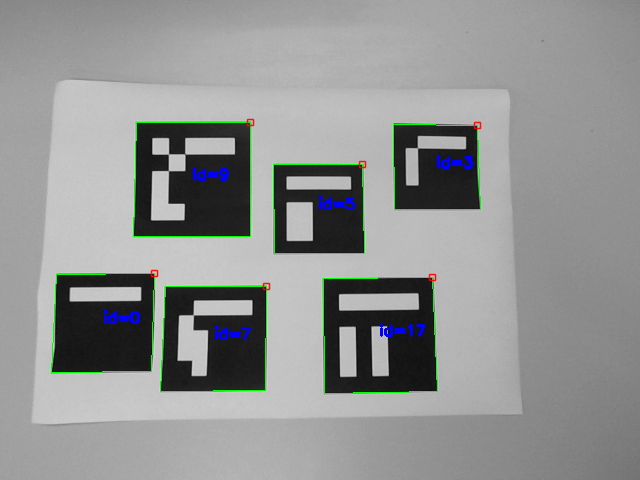}}
    \hfill
    \subfloat[Raw IR \label{fig:detections:raw_ir}]{%
        \includegraphics[width=0.45\linewidth]{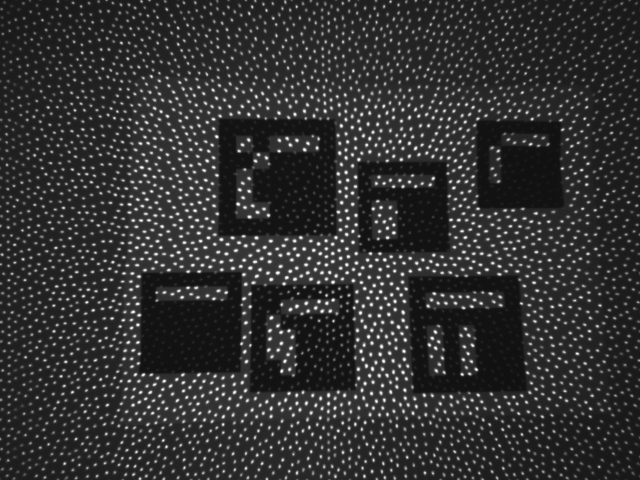}}
    \\
    \subfloat[CIDNet output\label{fig:detections:cid}]{%
        \includegraphics[width=0.45\linewidth]{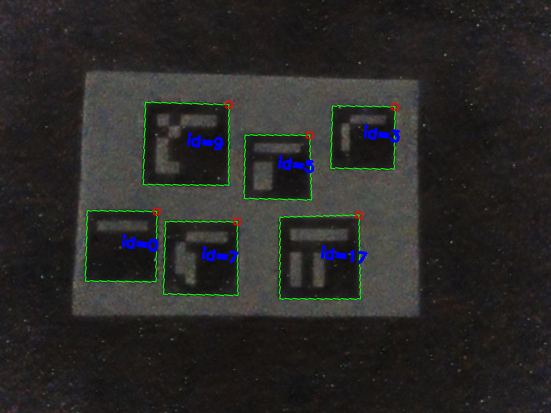}}
    \hfill
    \subfloat[DeepMAO (ablation)\label{fig:detections:deepmao_ir}]{%
        \includegraphics[width=0.45\linewidth]{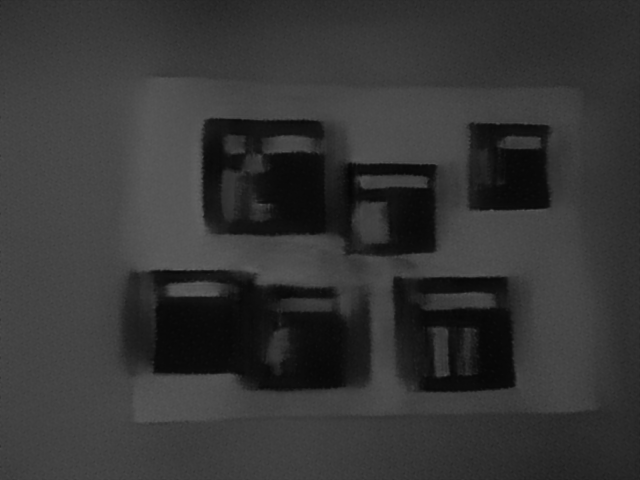}}
    \\
    \subfloat[U-Net output\label{fig:detections:unet_aruco}]{%
        \includegraphics[width=0.45\linewidth]{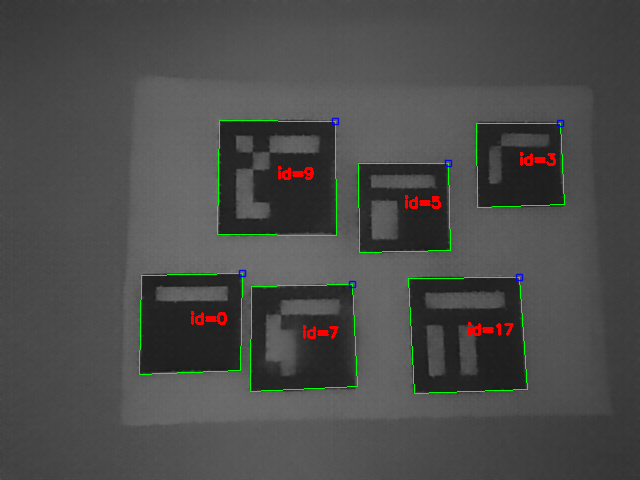}}
    \hfill
    \subfloat[CLEAR-IR output\label{fig:detections:denoised_ir}]{%
        \includegraphics[width=0.45\linewidth]{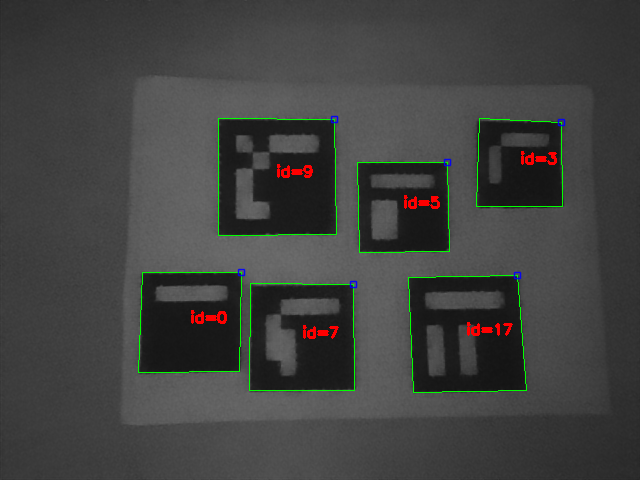}}
    \caption{ArUco marker detection. (a) RGB provides the reference; all 6 markers are detected. (b) Raw IR: 0/6 detected due to emitter dot corruption. (c) CIDNet: 6/6 detected (applied to passive dark RGB stream). (d) Original DeepMAO (retrained on IR dataset): 0/6 — over smoothing destroys marker grids. (e) U-Net baseline: 6/6. (f) CLEAR-IR: 6/6, with clean marker boundaries.}
    \label{fig:detections}
\end{figure}

ArUco markers are fiducial markers widely used in robotics for localisation and pose estimation. The ability to reliably detect these markers is crucial for robust navigation. However, structured light patterns can interfere with the marker's distinct black and white grid, often leading to detection failures. This interference is clearly visible in the raw infrared (IR) image shown in Figure \ref{fig:detections:raw_ir}. For comparison, a baseline for successful detection is provided by the standard RGB image, as seen in Figure \ref{fig:detections:rgb}.

The CLEAR-IR framework addresses these challenges by enabling robust ArUco marker detection in IR images. By removing the spurious dot patterns, the model ensures the marker's grid is clearly visible and interpretable by the detection algorithm, as demonstrated in Figure \ref{fig:detections:denoised_ir}. This is a significant advancement for low light scenes where raw IR images are typically unusable for this task. The reconstructed output allows for consistent and accurate marker identification, thus enabling reliable localisation and mapping. This outcome directly contrasts with findings in prior work, such as \cite{icinco19}, which concluded that raw IR images were unsuitable for low light tag detection without the use of retroreflective materials. Similarly, this approach simplifies methods like \cite{s21030717}, which propose using specialised tags for low light localisation, by requiring no special tags or materials. 

\subsection{Visual SLAM Performance}
\label{sec:slam}

\begin{table*}[b!]
\centering
\caption{Quantitative evaluation of SLAM trajectory accuracy (RMSE in meters) and inference latency across Low Light (LL) and Extreme Low Light (ELL) scenarios.}
\label{tab:combined_slam_performance}
\resizebox{\textwidth}{!}{
\begin{tabular}{lcccccccc}
\toprule
\textbf{Method} & \multicolumn{3}{c}{\textbf{Low Light (LL) - RMSE $\pm$ std}} & \multicolumn{3}{c}{\textbf{Extreme Low Light (ELL) - RMSE $\pm$ std}} & \multicolumn{2}{c}{\textbf{Inference (ms)}} \\
\cmidrule(lr){2-4} \cmidrule(lr){5-7} \cmidrule(lr){8-9}
& \textbf{FB} & \textbf{SQ1} & \textbf{SQ2} & \textbf{FB} & \textbf{SQ1} & \textbf{SQ2} & \textbf{Mean} & \textbf{std} \\
\midrule
\multicolumn{9}{l}{\textit{RGB based methods (dark RGB stream)}} \\
Retinex \cite{Cai_2017_ICCV} & \textbf{0.064$\pm$0.02} & 0.163$\pm$0.05 & 0.168$\pm$0.05 & DNI & DNI & DNI & 73.43 & 10.54 \\
CLAHE \cite{CLAHE_Srikanth} & 0.069$\pm$0.03 & 0.156$\pm$0.05 & 0.169$\pm$0.05 & DNI & DNI & DNI & \textbf{15.16} & \textbf{3.09} \\
Zero-DCE \cite{Zero-DCE} & 0.065$\pm$0.02 & \textbf{0.147$\pm$0.05} & 0.190$\pm$0.05 & DNI & DNI & DNI & 46.35 & 2.32 \\
LLFormer \cite{wang2023ultra} & 0.067$\pm$0.03 & 0.156$\pm$0.05 & 0.183$\pm$0.05 & DNI & DNI & DNI & 236.88 & 10.20 \\
CIDNet \cite{yan2025hvi} & 0.069$\pm$0.03 & 0.560$\pm$0.22 & 0.178$\pm$0.05 & DNI & DNI & DNI & 76.33 & 0.86 \\
\midrule
\multicolumn{9}{l}{\textit{IR based methods }} \\
Raw IR & DNI & DNI & DNI & DNI & DNI & DNI & DNI & DNI \\
U-Net & 0.072$\pm$0.03 & 0.161$\pm$0.06 & 0.164$\pm$0.06 & 0.097$\pm$0.05 & 0.167$\pm$0.05 & 0.168$\pm$0.06 & 49.93 & 3.42 \\
\textbf{CLEAR-IR (Ours)} & 0.070$\pm$0.03 & 0.164$\pm$0.06 & \textbf{0.137$\pm$0.07} & \textbf{0.092$\pm$0.04} & \textbf{0.159$\pm$0.05} & \textbf{0.159$\pm$0.12} & 46.75 & 3.48 \\
\bottomrule
\multicolumn{9}{l}{\small \textit{DNI: Did Not Initialise; Bold shows best results}}
\end{tabular}
}
\end{table*}

\begin{figure*}[t!]
    \centering
    \subfloat[SQ1 trajectories -- low light\label{fig:sq1_ll}]{%
       \includegraphics[width=0.35\linewidth]{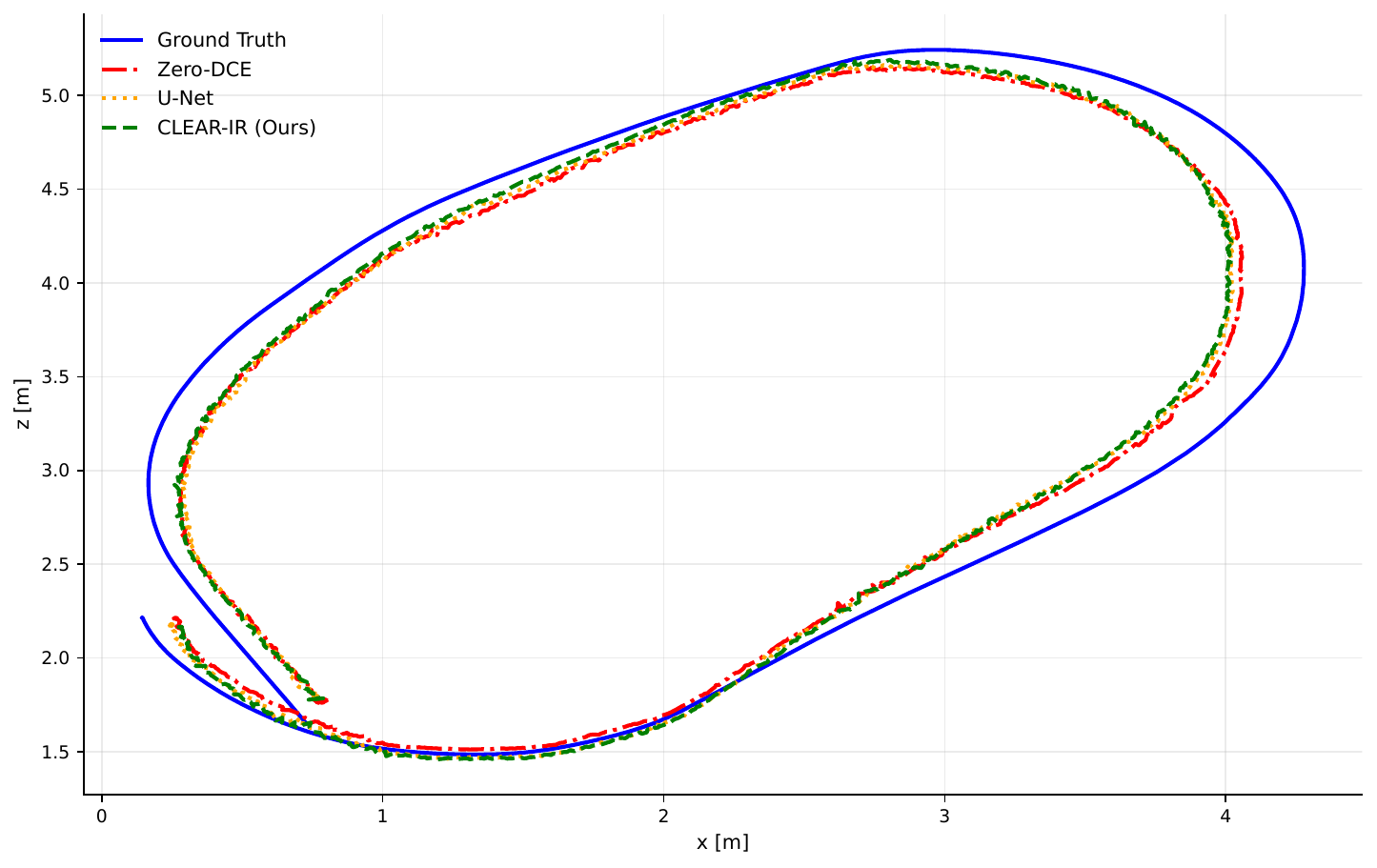}}
    \hspace{2em}
    \subfloat[SQ1 trajectories -- extreme low light\label{fig:sq1_ell}]{%
        \includegraphics[width=0.35\linewidth]{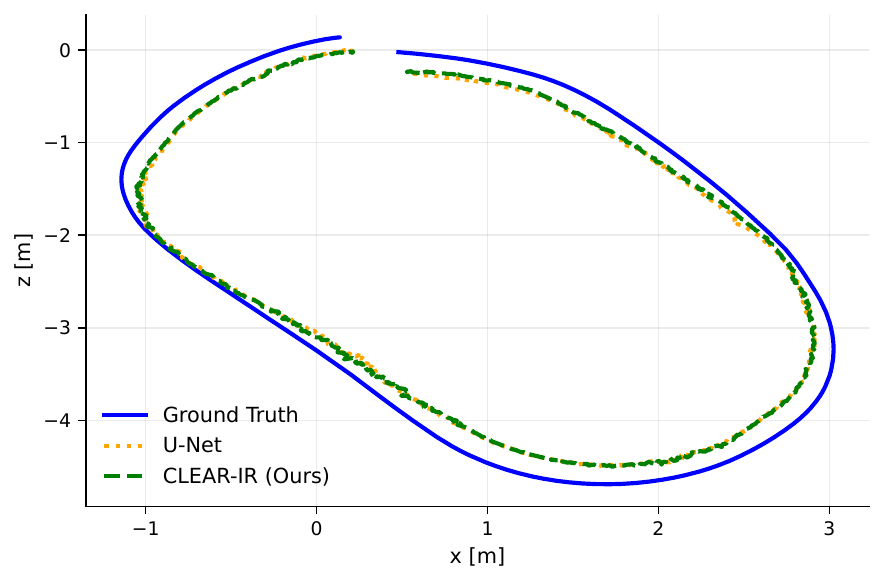}}
    \caption{SQ1 (Square 1 lap) trajectories. In low light (a), CLEAR-IR and U-Net closely follow the ground truth alongside the best RGB methods. In extreme low light (b), all RGB methods fail to initialise}
    \label{fig:vslam_trajectories}
\end{figure*}

To evaluate the practical utility of CLEAR-IR for robotics applications, we integrated the reconstructed infrared stream into a modified ORB-SLAM3 \cite{tukan2023orbslam3} Visual SLAM pipeline. In our implementation, the default ORB feature extractor was replaced with SuperPoint \cite{detone2018superpoint}, and feature matching was adapted accordingly to use floating point descriptor matching with L2 distance. This substitution was necessary as the CNN based reconstruction smooths the fine grained pixel gradients that traditional corner detectors such as ORB depend on, whereas learned extractors like SuperPoint remain capable of identifying geometrically consistent features despite this smoothing. The ORB extractor was evaluated but failed to yield trackable features on the reconstructed output. This modification ensures compatibility with learned feature descriptors while preserving the full SLAM pipeline, including tracking, local mapping, and loop closure. Robust SLAM performance critically depends on stable and repeatable feature tracking across frames. In raw IR images, however, the structured light emitter pattern introduces dense artificial high frequency responses. These phantom features are inconsistent across viewpoints and frames, leading to incorrect correspondences, tracking divergence, and eventual tracking failure. This behaviour is reflected in the Table~\ref{tab:combined_slam_performance} where Raw IR fails in all sequences.

To ensure fair comparison with RGB based enhancement methods, monocular RGB and depth sequences were recaptured under controlled lighting conditions. Three trajectories were recorded:

\begin{itemize}
    \item Forward Backward (FB): This sequence is a simple movement of moving in a straight line till an object and reversing backward
    \item Square 1 lap (SQ1): This sequence is a simple movement of rotating along an arena and returning to the start point as seen in \autoref{fig:vslam_trajectories}.
    \item Square 2 lap (SQ2): This sequence repeats the SQ1 trajectory for two full laps, providing a benchmark for observing the model's ability to facilitate successful loop closure and drift correction
\end{itemize}

Table~\ref{tab:combined_slam_performance} reports results in low light conditions where limited ambient illumination is still present. Under these conditions, the RGB enhancement methods (Retinex, CLAHE, Zero-DCE, LLFormer, CIDNet) perform reliably because the RGB images retain sufficient structural information for feature extraction and tracking.
CLEAR-IR and the U-Net baseline achieve performance comparable to these RGB based methods. In simpler trajectories (FB and SQ1), CLEAR-IR performs on par with the RGB enhancement models. Notably, in the SQ2 sequence, CLEAR-IR achieves the lowest RMSE among all evaluated methods. Since SQ2 stresses loop closure and long term consistency, this result indicates that CLEAR-IR reconstructs temporally stable structures that reduce drift accumulation over extended trajectories.
Importantly, unlike raw IR, the reconstructed outputs do not introduce false high frequency artefacts from the emitter pattern. Instead, the learned reconstruction produces geometrically meaningful and temporally consistent features suitable for SLAM.

Furthermore in extreme low light conditions the RGB images become severely underexposed, containing insufficient texture or gradient information for reliable feature detection. Consequently, all RGB based enhancement methods fail to maintain tracking across every sequence, as indicated by missing RMSE values.
In contrast, the IR images remain unaffected by ambient lighting conditions. While raw IR still fails due to emitter induced phantom features, both U-Net and CLEAR-IR successfully enable full SLAM operation across all sequences. CLEAR-IR consistently outperforms the U-Net baseline in terms of RMSE, achieving the best accuracy in FB, SQ1, and SQ2. 
This demonstrates a key advantage of the proposed framework: when RGB information collapses, CLEAR-IR leverages the invariant structural information present in the IR modality to reconstruct geometrically consistent representations that remain usable for feature based SLAM.

In terms of inference latency, CLEAR-IR maintains competitive runtime performance. All inference latency measurements reported in Table \ref{tab:combined_slam_performance} were obtained on an NVIDIA RTX 4080 GPU. While classical methods such as CLAHE are significantly faster, they do not provide robustness under extreme lighting degradation. Conversely, transformer based models such as LLFormer incur substantially higher computational cost. CLEAR-IR achieves a favourable trade off between localisation accuracy and inference speed, maintaining compatibility with real time SLAM update requirements.

\section{Conclusion}
Utilising active infrared cameras already integrated into many robotic platforms, CLEAR-IR effectively addresses the challenge of emitter pattern removal in active IR imagery. The dual stream architecture produces visually coherent, artefact free reconstructions compatible with existing RGB trained computer vision pipelines in complete darkness. Evaluation shows improvements in object detection, ArUco marker detection, and visual SLAM. In low light with little ambient illumination, CLEAR-IR performs on par with leading RGB enhancement methods, while in extreme low light where RGB sensing cannot receive usable signal, CLEAR-IR is the only approach that maintains reliable perception. The framework provides a foundation for autonomous systems operating in challenging lighting conditions.

\section{Future Direction}
Future work will focus on improving computational efficiency through quantisation and pruning for embedded deployment, and extending evaluation to scattering environments such as dust and smoke. Additional directions include fine-tuning downstream detectors on reconstructed IR imagery and integrating IR-guided RGB enhancement for systems operating with partial visible light information.

\section*{Acknowledgment}
RAICo contributed to this work by providing funding and a supportive research environment. Jongyun Kim assisted the authors with setup and data collection in a dark environment.

\bibliographystyle{IEEEtran}
\bibliography{references}

\end{document}